\documentclass[11pt]{article}
\usepackage{acl2016}
\usepackage{times}
\usepackage{latexsym}
\usepackage{bm}
\usepackage{multirow}
\usepackage{amsfonts}
\usepackage[boxed,linesnumbered,lined]{algorithm2e}
\usepackage{algpseudocode}
\usepackage{graphicx}
\usepackage{hyphenat}

\aclfinalcopy 

\def\argmax{\mathop{\rm argmax}}%
\def\argmin{\mathop{\rm argmin}}%

\title{Minimum Risk Training for Neural Machine Translation}

\author{Shiqi Shen$^\dagger$, Yong Cheng$^\#$,  Zhongjun He$^+$,  Wei He$^+$, Hua Wu$^+$, Maosong Sun$^\dagger$, Yang Liu$^\dagger$\thanks{\ \ Corresponding author: Yang Liu.}\\
	$^\dagger$State Key Laboratory of Intelligent Technology and Systems  \\
	Tsinghua National Laboratory for Information Science and Technology \\
	Department of Computer Science and Technology, Tsinghua University, Beijing, China \\
	$^\#$Institute for Interdisciplinary Information Sciences, Tsinghua University, Beijing, China  \\
	$^+$Baidu Inc., Beijing, China \\
	\small \{vicapple22, chengyong3001\}@gmail.com,  \{hezhongjun, hewei06, wu\_hua\}@baidu.com,
	\{sms, liuyang2011\}@tsinghua.edu.cn}

\date{}

\begin{document}

\maketitle

\begin{abstract}
  We propose minimum risk training for end-to-end neural machine translation. Unlike conventional maximum likelihood estimation, minimum risk training is capable of optimizing model parameters directly with respect to arbitrary evaluation metrics, which are not necessarily differentiable. Experiments show that our approach achieves significant improvements over maximum likelihood estimation on a state-of-the-art neural machine translation system across various languages pairs. Transparent to architectures, our approach can be applied to more neural networks and potentially benefit more NLP tasks.
\end{abstract}

\section{Introduction}
Recently, end-to-end neural machine translation (NMT) \cite{Kalchbrenner:13,Sutskever:14,Bahdanau:15} has attracted increasing attention from the community. Providing a new paradigm for machine translation, NMT aims at training a single, large neural network that directly transforms a source-language sentence to a target-language sentence without explicitly modeling latent structures (e.g., word alignment, phrase segmentation, phrase reordering, and SCFG derivation) that are vital in conventional statistical machine translation (SMT) \cite{Brown:93,Koehn:03,Chiang:05}.

Current NMT models are based on the {\em encoder-decoder} framework \cite{Cho:14,Sutskever:14}, with an encoder to read and encode a source-language sentence into a vector, from which a decoder generates a target-language sentence. While early efforts encode the input into a fixed-length vector, Bahdanau et al. \shortcite{Bahdanau:15} advocate the attention mechanism to dynamically generate a context vector for a target word being generated.

Although NMT models have achieved results on par with or better than conventional SMT, they still suffer from a major drawback: the models are optimized to maximize the likelihood of training data instead of evaluation metrics that actually quantify translation quality. Ranzato et al. \shortcite{Ranzato:15} indicate two drawbacks of {\em maximum likelihood estimation} (MLE) for NMT. First, the models are only exposed to the training distribution instead of model predictions. Second, the loss function is defined at the word level instead of the sentence level.

In this work, we introduce {\em minimum risk training} (MRT) for neural machine translation. The new training objective is to minimize the expected loss (i.e., risk) on the training data. MRT has the following advantages over MLE:
\begin{enumerate}
\item {\em Direct optimization with respect to evaluation metrics}: MRT introduces evaluation metrics as loss functions and aims to minimize expected loss on the training data.
\item {\em Applicable to arbitrary loss functions}: our approach allows arbitrary sentence-level loss functions, which are not necessarily differentiable.
\item {\em Transparent to architectures}: MRT does not assume the specific architectures of NMT and can be applied to any end-to-end NMT systems.
\end{enumerate}

While MRT has been widely used in conventional SMT \cite{Och:03,Smith:06,He:12} and deep learning based MT \cite{Gao:14}, to the best of our knowledge, this work is the first effort to introduce MRT into end-to-end NMT. Experiments on a variety of language pairs (Chinese-English, English-French, and English-German) show that MRT leads to significant improvements over MLE on a state-of-the-art NMT system \cite{Bahdanau:15}.

\section{Background}
Given a source sentence $\mathbf{x} = \mathbf{x}_1,\dots, \mathbf{x}_m,\dots, \mathbf{x}_{M}$ and a target sentence $\mathbf{y}=\mathbf{y}_1,\dots,\mathbf{y}_n,\dots, \mathbf{y}_{N}$, end-to-end NMT directly models the translation probability:
\begin{eqnarray}
P(\mathbf{y}|\mathbf{x}; \bm{\theta}) = \prod_{n=1}^{N}P(\mathbf{y}_n | \mathbf{x}, \mathbf{y}_{<n}; \bm{\theta}),
\end{eqnarray}
where $\bm{\theta}$ is a set of model parameters and $\mathbf{y}_{<n}=\mathbf{y}_1,\dots,\mathbf{y}_{n-1}$ is a partial translation.

Predicting the $n$-th target word can be modeled by using a recurrent neural network:
\begin{eqnarray}
P(\mathbf{y}_n|\mathbf{x}, \mathbf{y}_{<n};\bm{\theta}) \propto \exp \Big \{q(\mathbf{y}_{n-1}, \mathbf{z}_{n}, \mathbf{c}_n, \bm{\theta}) \Big \},
\end{eqnarray}
where $\mathbf{z}_n$ is the $n$-th hidden state on the target side, $\bm{c}_n$ is the context for generating the $n$-th target word, and $q(\cdot)$ is a non-linear function. Current NMT approaches differ in calculating $\mathbf{z}_n$ and $\mathbf{c}_n$ and defining $q(\cdot)$. Please refer to \cite{Sutskever:14,Bahdanau:15} for more details.

Given a set of training examples $D = \{ \langle \mathbf{x}^{(s)}, \mathbf{y}^{(s)} \rangle \}_{s=1}^{S}$, the standard training objective is to maximize the log-likelihood of the training data:
\begin{eqnarray}
\hat{\bm{\theta}}_{\mathrm{MLE}} = \argmax_{\bm{\theta}}\Big\{ \mathcal{L}(\bm{\theta}) \Big \},
\end{eqnarray}
where
\begin{eqnarray}
\mathcal{L}(\bm{\theta})= \sum_{s=1}^{S} \log P(\mathbf{y}^{(s)}|\mathbf{x}^{(s)}; \bm{\theta}) \quad \quad \ \ \ \ \ \ \ \ \\
= \sum_{s=1}^{S}\sum_{n=1}^{N^{(s)}}\log P(\mathbf{y}_n^{(s)}|\mathbf{x}^{(s)},\mathbf{y}^{(s)}_{<n};\bm{\theta}).
\end{eqnarray}
We use $N^{(s)}$ to denote the length of the $s$-th target sentence $\mathbf{y}^{(s)}$.

The partial derivative with respect to a model parameter $\bm{\theta}_i$ is calculated as
\begin{eqnarray}
\frac{\partial \mathcal{L}(\bm{\theta})}{\partial \bm{\theta}_i} = \sum_{s=1}^{S} \sum_{n=1}^{N^{(s)}} \frac{\partial P(\mathbf{y}^{(s)}_n|\mathbf{x}^{(s)}, \mathbf{y}^{(s)}_{<n}; \bm{\theta}) / \partial \bm{\theta}_i}{P(\mathbf{y}^{(s)}_n|\mathbf{x}^{(s)}, \mathbf{y}^{(s)}_{<n}; \bm{\theta})}. \label{eq:mle_d}
\end{eqnarray}

\begin{table*}
\centering
\begin{tabular}{c||c||r|r|r|r}
& $\Delta(\mathbf{y}, \mathbf{y}^{(s)})$ & \multicolumn{4}{c}{$P(\mathbf{y}|\mathbf{x}^{(s)}; \bm{\theta})$} \\
\hline
\hline
$\mathbf{y}_1$ & $-1.0$ & $0.2$ & $0.3$ & $0.5$ & $0.7$ \\
$\mathbf{y}_2$ & $-0.3$ & $0.5$ & $0.2$ & $0.2$ & $0.1$ \\
$\mathbf{y}_3$ & $-0.5$ & $0.3$ & $0.5$ & $0.3$ & $0.2$ \\
\hline \hline
\multicolumn{2}{c||}{$\mathbb{E}_{\mathbf{y}|\mathbf{x}^{(s)}; \bm{\theta}}[\Delta(\mathbf{y}, \mathbf{y}^{(s)})]$} & $-0.50$ & $-0.61$ & $-0.71$ & $-0.83$
\end{tabular}
\caption{Example of minimum risk training. $\mathbf{x}^{(s)}$ is an observed source sentence, $\mathbf{y}^{(s)}$ is its corresponding gold-standard translation, and $\mathbf{y}_1$, $\mathbf{y}_2$, and $\mathbf{y}_3$  are model predictions. For simplicity, we suppose that the full search space contains only  three candidates. The loss function $\Delta(\mathbf{y}, \mathbf{y}^{(s)})$ measures the difference between model prediction and gold-standard. The goal of MRT is to find a distribution (the last column) that correlates well with the gold-standard by minimizing the expected loss.} \label{table:risk_example}
\end{table*}

Ranzato et al. \shortcite{Ranzato:15} point out that MLE for end-to-end NMT suffers from two drawbacks. First, while the models are trained only on the training data distribution, they are used to generate target words on previous model predictions, which can be erroneous, at test time. This is referred to as {\em exposure bias} \cite{Ranzato:15}. Second, MLE usually uses the cross-entropy loss focusing on word-level errors to maximize the probability of the next correct word, which might hardly correlate well with corpus-level and sentence-level evaluation metrics such as BLEU \cite{Papineni:02} and TER \cite{Snover:06}.

As a result, it is important to introduce new training algorithms for end-to-end NMT to include model predictions during training and optimize model parameters directly with respect to evaluation metrics.

\section{Minimum Risk Training for Neural Machine Translation}

Minimum risk training (MRT), which aims to minimize the expected loss on the training data, has been widely used in conventional SMT \cite{Och:03,Smith:06,He:12} and deep learning based MT \cite{Gao:14}. The basic idea is to introduce evaluation metrics as loss functions and assume that the optimal set of model parameters should minimize the expected loss on the training data.

Let $\langle \mathbf{x}^{(s)}, \mathbf{y}^{(s)} \rangle$ be the $s$-th sentence pair in the training data and $\mathbf{y}$ be a model prediction. We use a {\em loss function} $\Delta(\mathbf{y}, \mathbf{y}^{(s)})$ to measure the discrepancy between the model prediction $\mathbf{y}$ and the gold-standard translation $\mathbf{y}^{(s)}$. Such a loss function can be negative smoothed sentence-level evaluation metrics such as BLEU \cite{Papineni:02}, NIST \cite{Doddington:02}, TER \cite{Snover:06}, or METEOR \cite{Lavie:09} that have been widely used in machine translation evaluation. Note that a loss function is not parameterized and thus not differentiable.

In MRT, the {\em risk} is defined as the expected loss with respect to the posterior distribution:
\begin{eqnarray}
\mathcal{R}(\bm{\theta}) = \sum_{s=1}^{S} \mathbb{E}_{\mathbf{y}|\mathbf{x}^{(s)}; \bm{\theta}}\Big[ \Delta(\mathbf{y}, \mathbf{y}^{(s)}) \Big] \ \ \ \ \ \ \ \ \ \ \ \ \ \ \ \ \ \ \ \\
= \sum_{s=1}^{S} \sum_{\mathbf{y} \in \mathcal{Y}(\mathbf{x}^{(s)})}P(\mathbf{y}|\mathbf{x}^{(s)}; \bm{\theta})\Delta(\mathbf{y}, \mathbf{y}^{(s)}),
\end{eqnarray}
where $\mathcal{Y}(\mathbf{x}^{(s)})$ is a set of all possible candidate translations for $\mathbf{x}^{(s)}$.

The training objective of MRT is to minimize the risk on the training data:
\begin{eqnarray}
\hat{\bm{\theta}}_{\mathrm{MRT}} = \argmin_{\bm{\theta}}\Big\{ \mathcal{R}(\bm{\theta}) \Big\}.
\end{eqnarray}

Intuitively, while MLE aims to maximize the likelihood of training data, our
training objective is to discriminate between candidates. For example, in Table \ref{table:risk_example}, suppose the candidate set $\mathcal{Y}(\mathbf{\mathbf{x}^{(s)}})$  contains only three candidates: $\mathbf{y}_1$, $\mathbf{y}_2$, and $\mathbf{y}_3$. According to the losses calculated by comparing with the gold-standard translation $\mathbf{y}^{(s)}$, it is clear that $\mathbf{y}_1$ is the best candidate, $\mathbf{y}_3$ is the second best, and $\mathbf{y}_2$ is the worst: $\mathbf{y}_1 > \mathbf{y}_3 > \mathbf{y}_2$. The right half of Table \ref{table:risk_example} shows four models. As model 1 (column 3) ranks the candidates in a reverse order as compared with the gold-standard (i.e., $\mathbf{y}_2 > \mathbf{y}_3 > \mathbf{y}_1$), it obtains the highest risk of $-0.50$. Achieving a better correlation with the gold-standard than model 1 by predicting $\mathbf{y}_3 > \mathbf{y}_1 > \mathbf{y}_2$, model 2 (column 4) reduces the risk to $-0.61$. As model 3 (column 5) ranks the candidates in the same order with the gold-standard, the risk goes down to $-0.71$. The risk can be further reduced by concentrating the probability mass on $\mathbf{y}_1$ (column 6). As a result, by minimizing the risk on the training data, we expect to obtain a model that correlates well with the gold-standard.

In MRT, the partial derivative with respect to a model parameter $\bm{\theta}_i$ is given by
\begin{eqnarray}
\frac{\partial \mathcal{R}(\bm{\theta})}{\partial \bm{\theta}_i} \quad \quad \quad \quad \quad \quad \quad \quad \quad \quad \quad \quad \quad \quad \quad \ \ \nonumber \\
= \sum_{s=1}^{S} \mathbb{E}_{\mathbf{y}|\mathbf{x}^{(s)}; \bm{\theta}}\Bigg[ \Delta(\mathbf{y}, \mathbf{y}^{(s)}) \times \quad \quad \quad \quad \quad \quad \quad \ \nonumber \\
\sum_{n=1}^{N^{(s)}} \frac{\partial P(\mathbf{y}_n|\mathbf{x}^{(s)}, \mathbf{y}_{<n}; \bm{\theta}) / \partial \bm{\theta}_i}{P(\mathbf{y}_n|\mathbf{x}^{(s)}, \mathbf{y}_{<n}; \bm{\theta})}  \Bigg]. \label{eq:mrt_d}
\end{eqnarray}

Since Eq. (\ref{eq:mrt_d}) suggests there is no need to differentiate $\Delta(\mathbf{y}, \mathbf{y}^{(s)})$, MRT allows arbitrary non-differentiable loss functions. In addition, our approach is transparent to architectures and can be applied to arbitrary end-to-end NMT models.

Despite these advantages, MRT faces a major challenge: the expectations in Eq. (\ref{eq:mrt_d}) are usually intractable to calculate due to the exponential search space of $\mathcal{Y}(\mathbf{x}^{(s)})$, the non-decomposability of the loss function $\Delta(\mathbf{y}, \mathbf{y}^{(s)})$, and the context sensitiveness of NMT.

To alleviate this problem, we propose to only use a subset of the full search space to approximate the posterior distribution and introduce a new training objective:
\begin{eqnarray}
\tilde{\mathcal{R}}(\bm{\theta}) = \sum_{s=1}^{S} \mathbb{E}_{\mathbf{y}|\mathbf{x}^{(s)}; \bm{\theta}, \alpha}\Big[ \Delta(\mathbf{y}, \mathbf{y}^{(s)}) \Big] \quad \quad \quad \quad \ \ \ \ \\
= \sum_{s=1}^{S} \sum_{\mathbf{y} \in \mathcal{S}(\mathbf{x}^{(s)})} Q(\mathbf{y}|\mathbf{x}^{(s)}; \bm{\theta}, \alpha) \Delta(\mathbf{y}, \mathbf{y}^{(s)}),
\end{eqnarray}
where $\mathcal{S}(\mathbf{x}^{(s)}) \subset \mathcal{Y}(\mathbf{x}^{(s)})$ is a sampled subset of the full search space, and $Q(\mathbf{y}|\mathbf{x}^{(s)}; \bm{\theta}, \alpha)$ is a distribution defined on the subspace $\mathcal{S}(\mathbf{x}^{(s)})$:
\begin{eqnarray}
Q(\mathbf{y}|\mathbf{x}^{(s)}; \bm{\theta}, \alpha) = \frac{P(\mathbf{y}|\mathbf{x}^{(s)}; \bm{\theta})^{\alpha}}{\sum_{\mathbf{y}' \in \mathcal{S}(\mathbf{x}^{(s)})} P(\mathbf{y}'|\mathbf{x}^{(s)};\bm{\theta})^{\alpha}}. \label{eq:Q}
\end{eqnarray}
Note that $\alpha$ is a hyper-parameter that controls the sharpness of the $Q$ distribution \cite{Och:03}.

\IncMargin{1em}
\begin{algorithm*}[!t]
\small
\caption{Sampling the full search space.} \label{alg}
\KwIn{the $s$-th source sentence in the training data $\mathbf{x}^{(s)}$, the $s$-th target sentence in the training data $\mathbf{y}^{(s)}$, the set of model parameters $\bm{\theta}$, the limit on the length of a candidate translation $l$, and the limit on the size of sampled space $k$.}
\KwOut{sampled space $\mathcal{S}(\mathbf{x}^{(s)}).$}
$\mathcal{S}(\mathbf{x}^{(s)}) \gets \{ \mathbf{y}^{(s)}\}$\tcp*[l]{the gold-standard translation is included}
$i \gets 1$; \\
\While{$i \le k$}
{
    $\mathbf{y} \gets \emptyset$\tcp*[l]{an empty candidate translation}
    $n \gets 1$\;
    \While {$n \le l$}
    {
        $y \sim P(\mathbf{y}_n|\mathbf{x}^{(s)}, \mathbf{y}_{<n}; \bm{\theta})$\tcp*[l]{sample the $n$-th target word}
        $\mathbf{y} \gets \mathbf{y} \cup \{y\}$\;
        \If{$y = \mathrm{EOS}$}
        {
            \textbf{break}\tcp*[l]{terminate if reach the end of sentence}
        }
        $n \gets n + 1$\;
    }

    $\mathcal{S}(\mathbf{x}^{(s)}) \gets \mathcal{S}(\mathbf{x}^{(s)}) \cup \{\mathbf{y}\}$\;

    $i \gets i + 1$;
}
\end{algorithm*}
\DecMargin{1em}

Algorithm \ref{alg} shows how to build $\mathcal{S}(\mathbf{x}^{(s)})$ by sampling the full search space. The sampled subset initializes with the gold-standard translation (line 1). Then, the algorithm keeps sampling a target word given the source sentence and the partial translation until reaching the end of sentence (lines 3-16). Note that sampling might produce duplicate candidates, which are removed when building the subspace. We find that it is inefficient to force the algorithm to generate exactly $k$ distinct candidates because high-probability candidates can be sampled repeatedly, especially when the probability mass highly concentrates on a few candidates. In practice, we take advantage of GPU's parallel architectures to speed up the sampling. \footnote{To build the subset, an alternative to sampling is computing top-$k$ translations. We prefer sampling to computing top-$k$ translations for efficiency: sampling is more efficient and easy-to-implement than calculating $k$-best lists, especially given the extremely parallel architectures of GPUs.}

Given the sampled space, the partial derivative with respect to a model parameter $\bm{\theta}_i$ of $\tilde{\mathcal{R}}(\bm{\theta})$ is given by
\begin{eqnarray}
\frac{\partial \tilde{R}(\bm{\theta})}{\partial \bm{\theta}_i} \quad \quad \quad \quad \quad \quad \quad \quad \quad \quad \quad \quad \quad \quad \ \ \nonumber \\
= \alpha \sum_{s=1}^{S} \mathbb{E}_{\mathbf{y}|\mathbf{x}^{(s)}; \bm{\theta}, \alpha}\Bigg[ \frac{\partial P(\mathbf{y}|\mathbf{x}^{(s)}; \bm{\theta})/\partial \bm{\theta}_i}{P(\mathbf{y}|\mathbf{x}^{(s)}; \bm{\theta})} \times \quad \ \ \ \nonumber \\
\Big( \Delta(\mathbf{y}, \mathbf{y}^{(s)}) - \nonumber \quad \quad \quad \quad \ \ \ \ \\
\quad \quad \quad \quad \quad \quad \quad \quad  \mathbb{E}_{\mathbf{y}'|\mathbf{x}^{(s)};\bm{\theta}, \alpha}\big[ \Delta(\mathbf{y}', \mathbf{y}^{(s)}) \big]\Big) \Bigg]. \label{eq:mrt_d2}
\end{eqnarray}

Since $|\mathcal{S}(\mathbf{x}^{(s)})| \ll |\mathcal{Y}(\mathbf{x}^{(s)})|$, the expectations in Eq. (\ref{eq:mrt_d2}) can be efficiently calculated by explicitly enumerating all candidates in $\mathcal{S}(\mathbf{x}^{(s)})$. In our experiments, we find that approximating the full space with $100$ samples (i.e., $k=100$) works very well and further increasing sample size does not lead to significant improvements (see Section 4.3).

\section{Experiments}

\subsection{Setup}
We evaluated our approach on three translation tasks: Chinese-English, English-French, and English-German. The evaluation metric is BLEU \cite{Papineni:02} as calculated by the \verb|multi-bleu.perl| script.

For Chinese-English, the training data consists of 2.56M pairs of sentences with 67.5M Chinese words and 74.8M English words, respectively. We used the NIST 2006 dataset as the validation set (hyper-parameter optimization and model selection) and the NIST 2002, 2003, 2004, 2005, and 2008 datasets as test sets.

For English-French, to compare with the results reported by previous work on end-to-end NMT \cite{Sutskever:14,Bahdanau:15,Jean:15,Luong:15}, we used the same subset of the WMT 2014 training corpus that contains 12M sentence pairs with 304M English words and 348M French words. The concatenation of news-test 2012 and news-test 2013 serves as the validation set and news-test 2014 as the test set.

For English-German, to compare with the results reported by previous work \cite{Jean:15,Luong:15a}, we used the same subset of the WMT 2014 training corpus that contains 4M sentence pairs with 91M English words and 87M German words. The concatenation of news-test 2012 and news-test 2013 is used as the validation set and news-test 2014 as the test set.

\begin{figure}[!t]
	\begin{center}
		\includegraphics[width=0.5\textwidth]{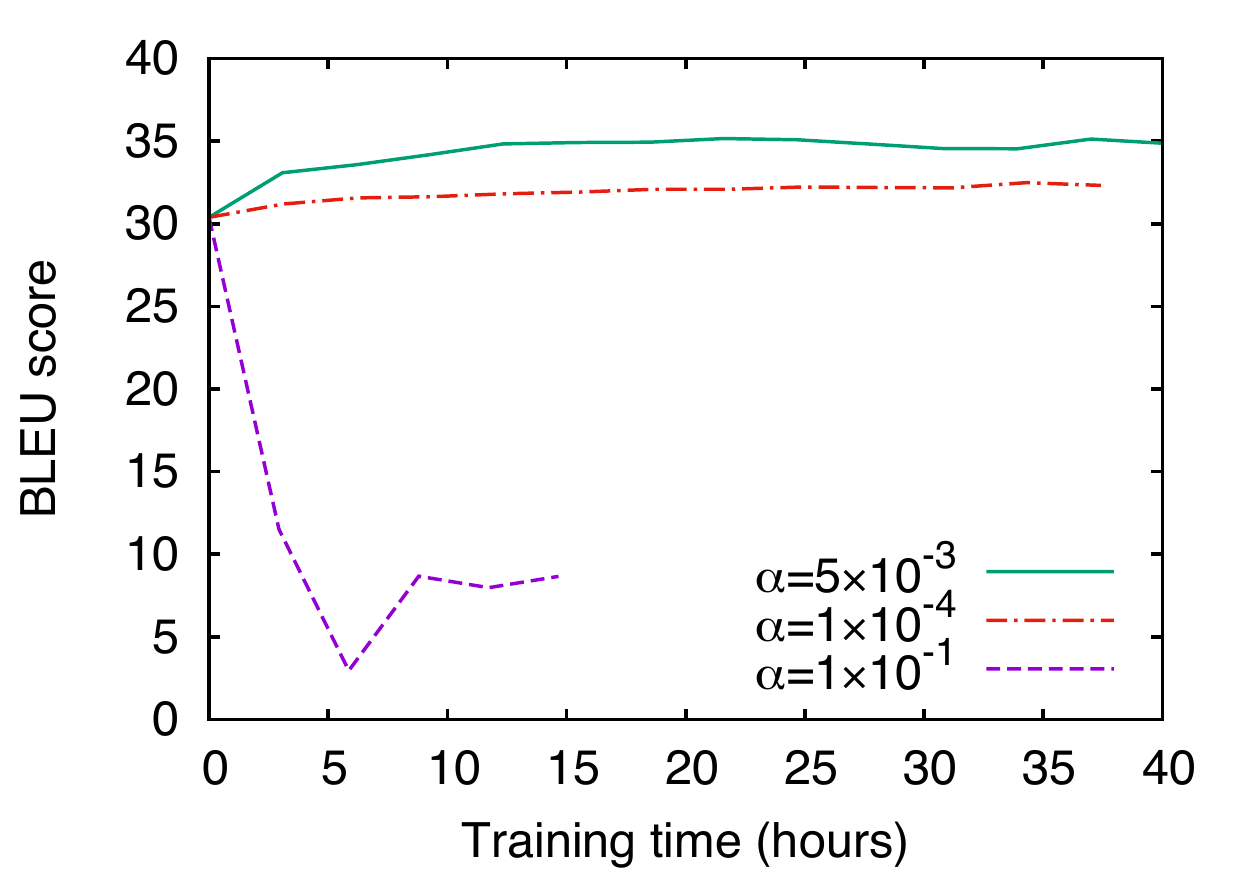}
	\end{center}
	\caption{Effect of $\alpha$ on the Chinese-English validation set. }
	\label{fig:effect_alpha}
\end{figure}

We compare our approach with two state-of-the-art SMT and NMT systems:

\begin{enumerate}
\item \textproc{Moses} \cite{Koehn:07}: a phrase-based SMT system using minimum error rate training \cite{Och:03}.

\item \textproc{RNNsearch} \cite{Bahdanau:15}: an attention-based NMT system using maximum likelihood estimation.
\end{enumerate}

\textproc{Moses} uses the parallel corpus to train a phrase-based translation model and the target part to train a 4-gram language model using the SRILM toolkit \cite{Stolcke:02}. \footnote{It is possible to exploit larger monolingual corpora for both \textproc{Moses} and \textproc{RNNsearch} \cite{Gulcehre:15,Sennrich:15}. We leave this for future work.} The log-linear model Moses uses is trained by the minimum error rate training (MERT) algorithm \cite{Och:03} that directly optimizes model parameters with respect to evaluation metrics.

\textproc{RNNsearch} uses the parallel corpus to train an attention-based neural translation model using the maximum likelihood criterion.

\begin{figure}[!t]
	\begin{center}
		\includegraphics[width=0.5\textwidth]{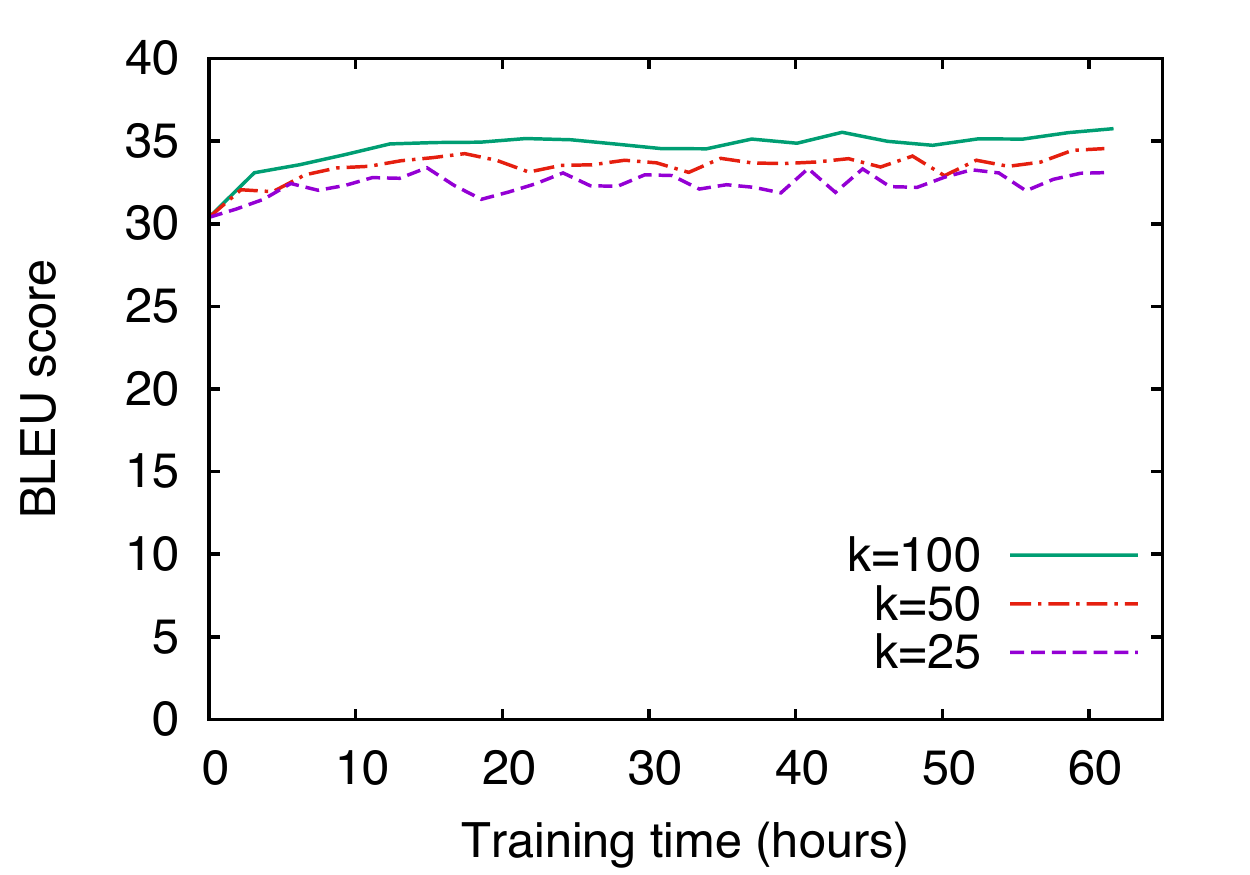}
	\end{center}
	\caption{Effect of sample size on the Chinese-English validation set. }
	\label{fig:effect_sample}
\end{figure}

On top of \textproc{RNNsearch}, our approach replaces MLE with MRT. We initialize our model with the RNNsearch50 model \cite{Bahdanau:15}.  We set the vocabulary size to 30K for Chinese-English and English-French and 50K for English-German. The beam size for decoding is 10. The default loss function is negative smoothed sentence-level BLEU.

\subsection{Effect of $\alpha$}

The hyper-parameter $\alpha$ controls the smoothness of the $Q$ distribution (see Eq. (\ref{eq:Q})). As shown in Figure \ref{fig:effect_alpha}, we find that $\alpha$ has a critical effect on BLEU scores on the Chinese-English validation set. While $\alpha = 1\times 10^{-1}$ deceases BLEU scores dramatically, $\alpha = 5\times 10^{-3}$ improves translation quality significantly and consistently. Reducing $\alpha$ further to $1\times 10^{-4}$, however, results in lower BLEU scores. Therefore, we set $\alpha = 5 \times 10^{-3}$ in the following experiments.

\begin{table}[!t]
\begin{tabular}{c|c||c|c|c}
criterion & loss & BLEU  & TER & NIST \\
\hline \hline
MLE & N/A & 30.48 & 60.85 & 8.26 \\
\hline
 & $-$sBLEU & {\bf 36.71} & {\bf 53.48} &  {\bf 8.90} \\
MRT & sTER & 30.14 & 53.83 & 6.02 \\
 & $-$sNIST & 32.32 & 56.85 & {\bf 8.90}
\end{tabular}
\caption{Effect of loss function on the Chinese-English validation set. } \label{table:loss}
\end{table}

\subsection{Effect of Sample Size}

For efficiency, we sample $k$ candidate translations from the full search space $\mathcal{Y}(\mathbf{x}^{(s)})$ to build an approximate posterior distribution $Q$ (Section 3). Figure \ref{fig:effect_sample} shows the effect of sample size $k$ on the Chinese-English validation set. It is clear that BLEU scores consistently rise with the increase of $k$. However, we find that a sample size larger than 100 (e.g., $k = 200$) usually does not lead to significant improvements and increases the GPU memory requirement. Therefore, we set $k = 100$ in the following experiments.

\subsection{Effect of Loss Function}

As our approach is capable of incorporating evaluation metrics as loss functions, we investigate the effect of different loss functions on BLEU, TER and NIST scores on the Chinese-English validation set. As shown in Table \ref{table:loss}, negative smoothed sentence-level BLEU (i.e, $-$sBLEU) leads to statistically significant improvements over MLE ($p<0.01$).
Note that the loss functions are all defined at the sentence level while evaluation metrics are calculated at the corpus level. This discrepancy might explain why optimizing with respect to sTER does not result in the lowest TER on the validation set.  As $-$sBLEU consistently improves all evaluation metrics, we use it as the default loss function in our experiments.

\subsection{Comparison of Training Time}

\begin{figure}[!t]
	\begin{center}
		\includegraphics[width=0.5\textwidth]{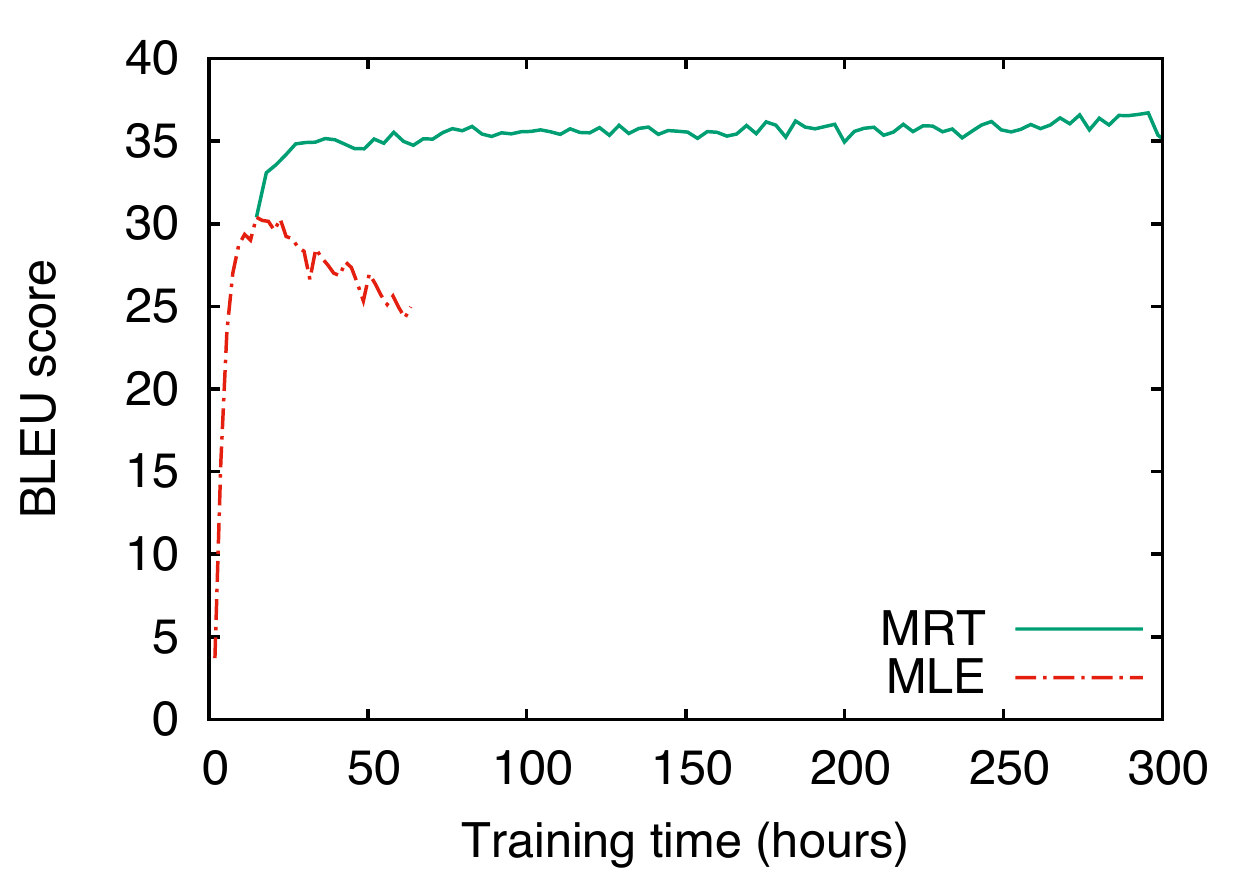}
	\end{center}
	\caption{Comparison of training time on the Chinese-English validation set. }
	\label{fig:time}
\end{figure}

We used a cluster with 20 Telsa K40 GPUs to train the NMT model. For MLE, it takes the cluster about one hour to train 20,000 mini-batches, each of which contains 80 sentences. The training time for MRT is longer than MLE: 13,000 mini-batches can be processed in one hour on the same cluster.

Figure \ref{fig:time} shows the learning curves of MLE and MRT on the validation set. For MLE, the BLEU score reaches its peak after about 20 hours and then keeps going down dramatically. Initializing with the best MLE model, MRT increases BLEU scores dramatically within about 30 hours. \footnote{Although it is possible to initialize with a randomized model, it takes much longer time to converge.} Afterwards, the BLEU score keeps improving gradually but there are slight oscillations.

\subsection{Results on Chinese-English Translation}

\begin{table*}[!t]
\centering
\begin{tabular}{c|c||c|ccccc}
System & Training & MT06 & MT02 & MT03 & MT04 & MT05 & MT08 \\
\hline \hline
\textproc{Moses} & MERT & 32.74 & 32.49 & 32.40 & 33.38 & 30.20 & 25.28 \\
\hline
\multirow{2}{*}{\textproc{RNNsearch}} & MLE & 30.70 & 35.13 & 33.73 & 	34.58 & 31.76 & 23.57 \\
\cline{2-8}
 & MRT & 37.34 & 40.36 & 40.93 & 41.37 & 38.81 & 29.23
\end{tabular}
\caption{Case-insensitive BLEU scores on Chinese-English translation. } \label{table:CE_BLEU}
\end{table*}

\begin{table*}[!t]
\centering
\begin{tabular}{c|c||c|ccccc}
System & Training & MT06 & MT02 & MT03 & MT04 & MT05 & MT08 \\
\hline \hline
\textproc{Moses} & MERT & 59.22 & 62.97 & 62.44 & 61.20 & 63.44 & 62.36 \\
\hline
\multirow{2}{*}{\textproc{RNNsearch}} & MLE & 60.74 & 58.94 & 60.10 & 58.91 & 61.74 & 64.52 \\
\cline{2-8}
 & MRT & 52.86 & 52.87 & 52.17 & 51.49 & 53.42 & 57.21
\end{tabular}
\caption{Case-insensitive TER scores on Chinese-English translation. } \label{table:CE_TER}
\end{table*}

\begin{figure}[!t]
	\begin{center}
		\includegraphics[width=0.5\textwidth]{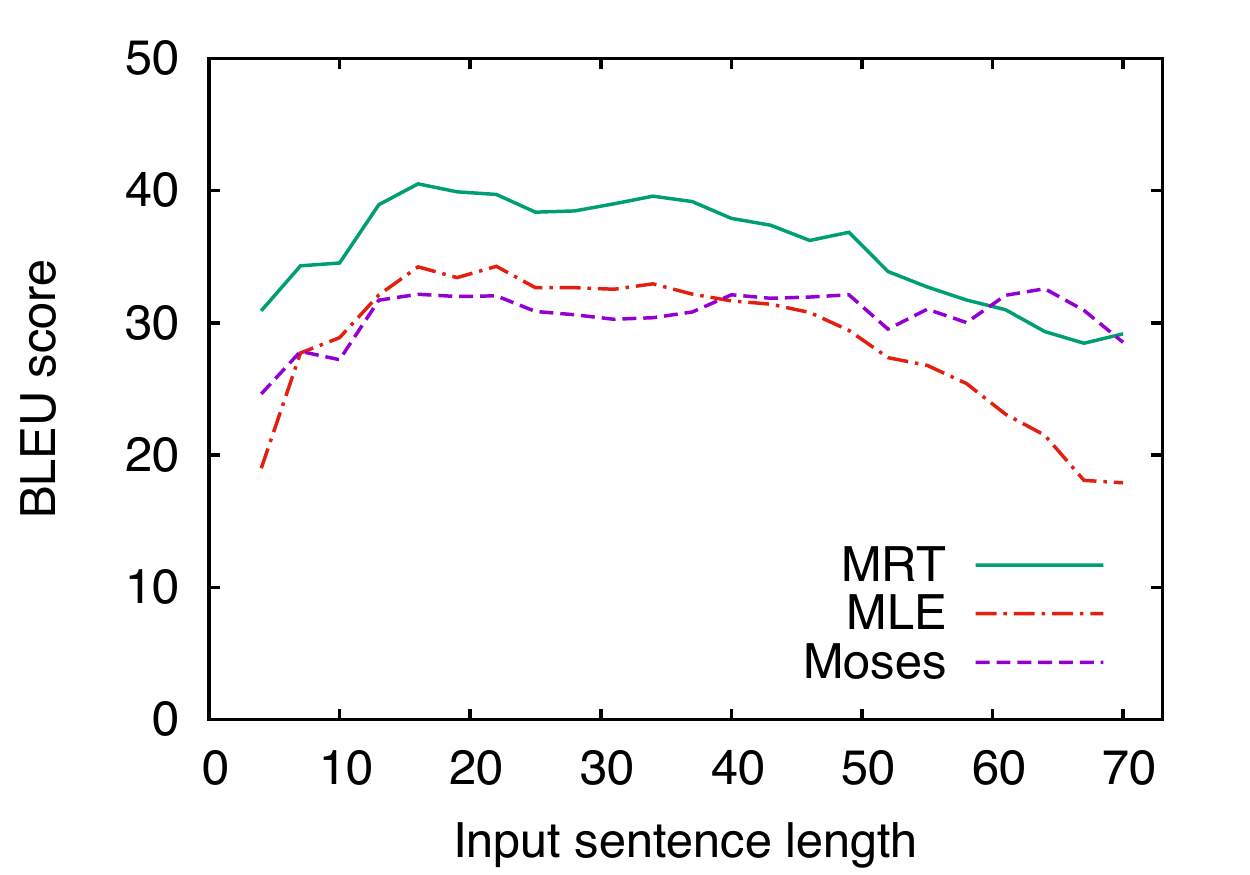}
	\end{center}
	\caption{BLEU scores on the Chinese-English test set over various input sentence lengths. } \label{fig:length_bleu}
\end{figure}

\subsubsection{Comparison of BLEU Scores}

Table \ref{table:CE_BLEU} shows BLEU scores on Chinese-English datasets. For \textproc{RNNsearch}, we follow Luong et al. \shortcite{Luong:15} to handle rare words. We find that introducing minimum risk training into neural MT leads to surprisingly substantial improvements over \textproc{Moses} and \textproc{RNNsearch}  with MLE as the training criterion (up to +8.61 and +7.20 BLEU points, respectively) across all test sets. All the improvements are statistically significant.

\subsubsection{Comparison of TER Scores}

Table \ref{table:CE_TER} gives TER scores on Chinese-English datasets. The loss function used in MRT is $-$sBLEU. MRT still obtains dramatic improvements over \textproc{Moses} and \textproc{RNNsearch}  with MLE as the training criterion (up to -10.27 and -8.32 TER points, respectively) across all test sets. All the improvements are statistically significant.

\begin{figure}[!t]
	\begin{center}
		\includegraphics[width=0.5\textwidth]{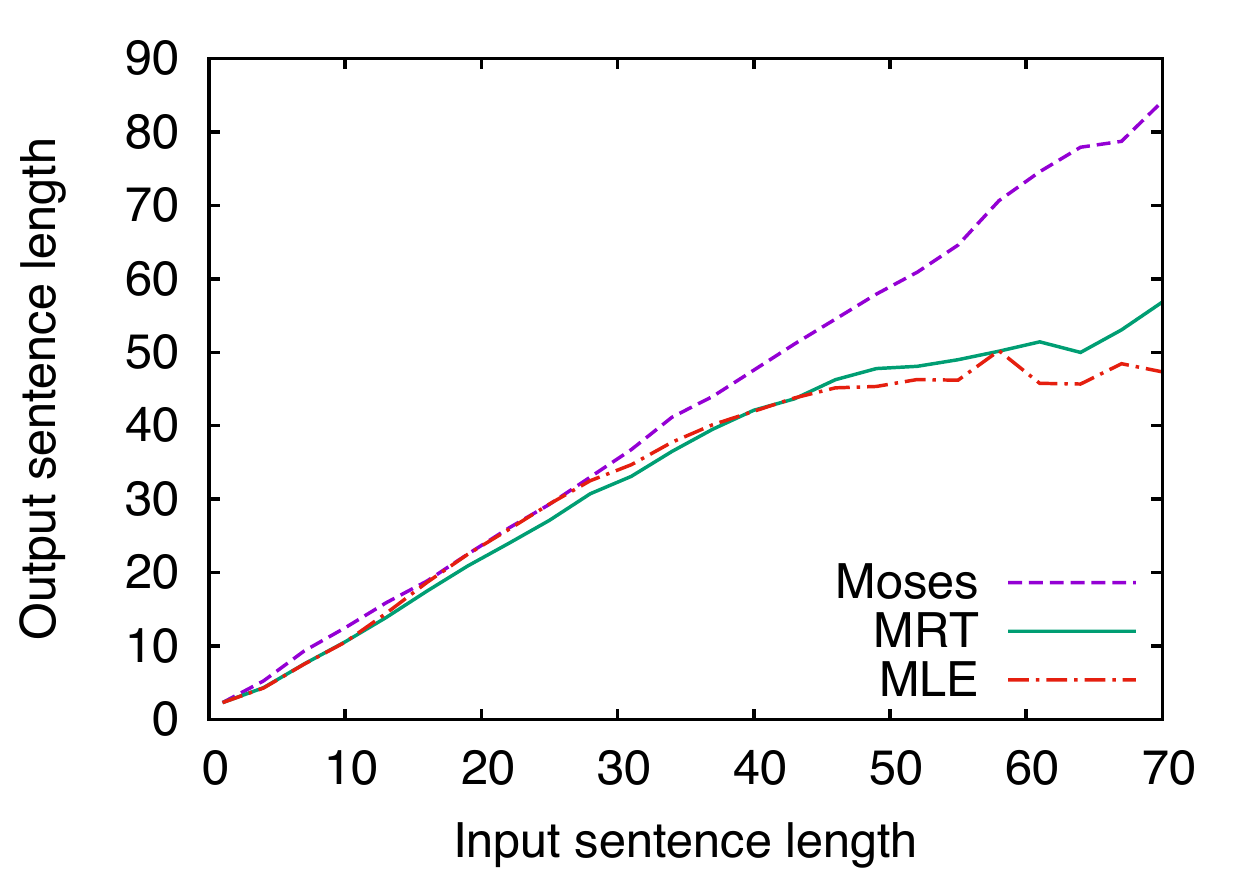}
	\end{center}
	\caption{Comparison of output sentences lengths on the Chinese-English test set. } \label{fig:length_length}
\end{figure}

\subsubsection{BLEU Scores over Sentence Lengths}

Figure \ref{fig:length_bleu} shows the BLEU scores of translations generated by \textproc{Moses}, \textproc{RNNsearch} with MLE, and \textproc{RNNsearch} with MRT on the Chinese-English test set with respect to input sentence lengths. While MRT consistently improves over MLE for all lengths, it achieves worse translation performance for sentences longer than 60 words.

One reason is that \textproc{RNNsearch} tends to produce short translations for long sentences. As shown in Figure \ref{fig:length_length}, both MLE and MRE generate much shorter translations than \textproc{Moses}. This results from the length limit imposed by \textproc{RNNsearch} for efficiency reasons: a sentence in the training set is no longer than 50 words. This limit deteriorates translation performance because the sentences in the test set are usually longer than 50 words.

\begin{table}[!t]
\centering
\begin{tabular}{c||c|c|c}
& \multicolumn{3}{c}{MLE vs. MRT} \\
\cline{2-4}
& $<$ & $=$ & $>$ \\
\hline \hline
evaluator 1 & 54\% & 24\% & 22\% \\
\hline
evaluator 2 & 53\% & 22\% & 25\%
\end{tabular}
\caption{Subjective evaluation of MLE and MRT on Chinese-English translation.} \label{table:human_eval}
\end{table}

\begin{table*}[!t]
\centering
\begin{tabular}{l|p{1.5\columnwidth}}
\hline
Source & {\em meiguo daibiao tuan baokuo laizi shidanfu daxue de yi wei zhongguo zhuanjia , liang ming canyuan waijiao zhengce zhuli yiji yi wei fuze yu pingrang dangju da jiaodao de qian guowuyuan guanyuan .} \\
\hline
Reference & the us delegation consists of a chinese expert from the stanford university , two senate foreign affairs policy assistants and a former state department official who was in charge of dealing with pyongyang authority . \\
\hline
\textproc{Moses} & the united states to members of the delegation include representatives from the stanford university , a chinese expert , two assistant senate foreign policy and a responsible for dealing with pyongyang before the officials of the state council . \\
\hline
\textproc{RNNsearch-MLE} & the us delegation comprises a chinese expert from stanford university , a chinese foreign office assistant policy assistant and a former official who is responsible for dealing with the pyongyang authorities . \\
\hline
\textproc{RNNsearch-MRT} & the us delegation included a chinese expert from the stanford university , two senate foreign policy assistants , and a former state department official who had dealings with the pyongyang authorities . \\
\hline
\end{tabular}
\caption{Example Chinese-English translations. ``Source'' is a romanized Chinese sentence, ``Reference'' is a gold-standard translation. ``\textproc{Moses}'' and ``\textproc{RNNsearch-MLE}'' are baseline SMT and NMT systems.  ``\textproc{RNNsearch-MRT}'' is our system.} \label{table:example}
\end{table*}

\begin{table*}[!t]
\centering
\begin{tabular}{l | l | c | c | c}
System & Architecture & Training & Vocab & BLEU \\
\hline \hline
\multicolumn{4}{c}{{\em Existing end-to-end NMT systems}} \\
\hline
Bahdanau et al. \shortcite{Bahdanau:15} & gated RNN with search & \multirow{8}{*}{MLE} & 30K & 28.45 \\
Jean et al. \shortcite{Jean:15} & gated RNN with search  & & 30K & 29.97 \\
Jean et al. \shortcite{Jean:15} & gated RNN with search + PosUnk & & 30K & 33.08 \\
Luong et al. \shortcite{Luong:15} & LSTM with 4 layers & & 40K & 29.50 \\
Luong et al. \shortcite{Luong:15} & LSTM with 4 layers + PosUnk & & 40K & 31.80 \\
Luong et al. \shortcite{Luong:15} & LSTM with 6 layers & & 40K & 30.40 \\
Luong et al. \shortcite{Luong:15} & LSTM with 6 layers + PosUnk  & & 40K & 32.70 \\
Sutskever et al. \shortcite{Sutskever:14} & LSTM with 4 layers & & 80K & 30.59 \\
\hline
\multicolumn{4}{c}{{\em Our end-to-end NMT systems}} \\
\hline
\multirow{3}{*}{{\em this work}} & gated RNN with search & MLE & 30K & 29.88 \\
& gated RNN with search & MRT & 30K & 31.30 \\
& gated RNN with search + PosUnk & MRT & 30K & 34.23 \\\end{tabular}
\caption{Comparison with previous work on English-French translation. The BLEU scores are case-sensitive. ``PosUnk'' denotes Luong et al. \shortcite{Luong:15}'s technique of handling rare words.} \label{table:EF}
\end{table*}

\begin{table*}[!t]
	\centering
	\begin{tabular}{l | l | c | c}
		System & Architecture & Training & BLEU \\
		\hline \hline
		\multicolumn{4}{c}{{\em Existing end-to-end NMT systems}} \\
		\hline
		Jean et al. \shortcite{Jean:15} & gated RNN with search  &\multirow{4}{*}{MLE} &  16.46 \\
		 Jean et al. \shortcite{Jean:15} & gated RNN with search + PosUnk & &  18.97 \\
		 Jean et al. \shortcite{Jean:15} & gated RNN with search + LV + PosUnk & &  19.40 \\
		 Luong et al. \shortcite{Luong:15a} & LSTM with 4 layers + dropout + local att. + PosUnk & &  20.90 \\
		\hline
		\multicolumn{4}{c}{{\em Our end-to-end NMT systems}} \\
		\hline
		\multirow{3}{*}{{\em this work}} & gated RNN with search & MLE &  16.45 \\
		& gated RNN with search & MRT & 18.02 \\
		& gated RNN with search + PosUnk & MRT & 20.45
	\end{tabular}
	\caption{Comparison with previous work on English-German translation. The BLEU scores are case-sensitive.} \label{table:EG}
\end{table*}

\subsubsection{Subjective Evaluation}

We also conducted a subjective evaluation to validate the benefit of replacing MLE with MRT. Two human evaluators were asked to compare MLE and MRT translations of 100 source sentences randomly sampled from the test sets without knowing from which system a candidate translation was generated.

Table \ref{table:human_eval} shows the results of subjective evaluation. The two human evaluators made close judgements: around 54\% of MLE translations are worse than MRE, 23\% are equal, and 23\% are better.

\subsubsection{Example Translations}

Table \ref{table:example} shows some example translations. We find that \textproc{Moses} translates a Chinese string ``{\em yi wei fuze yu pingrang dangju da jiaodao de qian guowuyuan guanyuan}'' that requires long-distance reordering in a wrong way, which is a notorious challenge for statistical machine translation. In contrast, \textproc{RNNsearch-MLE} seems to overcome this problem in this example thanks to the capability of gated RNNs to capture long-distance dependencies. However, as MLE uses a loss function defined only at the word level, its translation lacks sentence-level consistency: ``chinese'' occurs twice while ``two senate'' is missing. By optimizing model parameters directly with respect to sentence-level BLEU, \textproc{RNNsearch-MRT} seems to be able to generate translations more consistently at the sentence level.

\subsection{Results on English-French Translation}

Table \ref{table:EF} shows the results on English-French translation. We list existing end-to-end NMT systems that are comparable to our system.  All these systems use the same subset of the WMT 2014 training corpus and adopt MLE as the training criterion. They differ in network architectures and vocabulary sizes. Our \textproc{RNNsearch-MLE} system achieves a BLEU score comparable to that of Jean et al. \shortcite{Jean:15}. \textproc{RNNsearch-MRT} achieves the highest BLEU score in this setting even with a vocabulary size smaller  than Luong et al. \shortcite{Luong:15} and Sutskever et al. \shortcite{Sutskever:14}. Note that our approach does not assume specific architectures and can in principle be applied to any NMT systems.

\subsection{Results on English-German Translation}

Table \ref{table:EG} shows the results on English-German translation. Our approach still significantly outperforms MLE and achieves comparable results with state-of-the-art systems even though Luong et al. \shortcite{Luong:15a} used a much deeper neural network. We believe that our work can be applied to their architecture easily.

Despite these significant improvements, the margins on English-German and English-French datasets are much smaller than Chinese-English. We conjecture that there are two possible reasons. First, the Chinese-English datasets contain four reference translations for each sentence while both English-French and English-German datasets only have single references. Second, Chinese and English are more distantly related than English, French and German and thus benefit more from MRT that incorporates evaluation metrics into optimization to capture structural divergence.

\section{Related Work}

Our work originated from the minimum risk training algorithms in conventional statistical machine translation  \cite{Och:03,Smith:06,He:12}. Och \shortcite{Och:03} describes a smoothed error count to allow calculating gradients, which directly inspires us to use a parameter $\alpha$ to adjust the smoothness of the objective function. As neural networks are non-linear, our approach has to minimize the expected loss on the sentence level rather than the loss of 1-best translations on the corpus level. Smith and Eisner \shortcite{Smith:06} introduce minimum risk annealing for training log-linear models that is capable of gradually annealing to focus on the 1-best hypothesis. He et al. \shortcite{He:12} apply minimum risk training to learning phrase translation probabilities. Gao et al. \shortcite{Gao:14} leverage MRT for learning continuous phrase representations for statistical machine translation. The difference is that they use MRT to optimize a sub-model of SMT while we are interested in directly optimizing end-to-end neural translation models.

The Mixed Incremental Cross-Entropy Reinforce (MIXER) algorithm \cite{Ranzato:15} is in spirit closest to our work. Building on the REINFORCE algorithm proposed by Williams \shortcite{Williams:92}, MIXER allows incremental learning and the use of hybrid loss function that combines both REINFORCE and cross-entropy. The major difference is that Ranzato et al. \shortcite{Ranzato:15} leverage reinforcement learning while our work resorts to  minimum risk training. In addition, MIXER only samples one candidate to calculate reinforcement reward while MRT generates multiple samples to calculate the expected risk. Figure \ref{fig:effect_sample} indicates that multiple samples potentially increases MRT's capability of discriminating between diverse candidates and thus benefit translation quality. Our experiments confirm Ranzato et al. \shortcite{Ranzato:15}'s finding that taking evaluation metrics into account when optimizing model parameters does help to improve sentence-level text generation.

More recently, our approach has been successfully applied to summarization \cite{Ayana:16}. They optimize neural networks for headline generation with respect to ROUGE \cite{Lin:04} and also achieve significant improvements, confirming the effectiveness and applicability of our approach.

\section{Conclusion}
In this paper, we have presented a framework for minimum risk training in end-to-end neural machine translation. The basic idea is to minimize the expected loss in terms of evaluation metrics on the training data. We sample the full search space to approximate the posterior distribution to improve efficiency. Experiments show that MRT leads to significant improvements over maximum likelihood estimation for neural machine translation, especially for distantly-related languages such as Chinese and English.

In the future, we plan to test our approach on more language pairs and more end-to-end neural MT systems. It is also interesting to extend minimum risk training to minimum risk annealing following Smith and Eisner \shortcite{Smith:06}. As our approach is transparent to loss functions and architectures, we believe that it will also benefit more end-to-end neural architectures for other NLP tasks.

\section*{Acknowledgments}
This work was done while Shiqi Shen and Yong Cheng were visiting Baidu. Maosong Sun and Hua Wu are supported by the 973 Program (2014CB340501 \& 2014CB34505). Yang Liu is supported by the National Natural Science Foundation of China (No.61522204 and No.61432013) and the 863 Program (2015AA011808). This research is also supported by the Singapore National Research Foundation under its International Research Centre@Singapore Funding Initiative and administered by the IDM Programme.

\bibliography{acl2016_mrt}
\bibliographystyle{acl2016}

\end{document}